\documentclass[runningheads]{llncs}
\usepackage{graphicx}
\usepackage{amsmath,amssymb} 
\usepackage{gensymb} 
\usepackage{color}
\usepackage[width=122mm,left=12mm,paperwidth=146mm,height=193mm,top=12mm,paperheight=217mm]{geometry}

\usepackage{subcaption}
\usepackage{booktabs}
\usepackage{colortbl}
\RequirePackage{enumitem}
\setlist[itemize]{nosep}
\newcommand*{\MinNumber}{0}
\newcommand*{\MidNumber}{10} 
\newcommand*{\MaxNumber}{30}
\newcommand*{\Ratio}{70}
\usepackage{pgfplots}
\usepgfplotslibrary{dateplot}
\newcommand{\B}{\textbf}
\newcommand{\cl}[1]{%
        \ifdim #1 pt > \MidNumber pt
            \pgfmathsetmacro{\PercentColor}{max(min(\Ratio*(#1 - \MidNumber)/(\MaxNumber-\MidNumber),\Ratio),0.00)} %
            \edef\x{\noexpand\cellcolor{red!\PercentColor!yellow!80}}\x #1
        \else
            \pgfmathsetmacro{\PercentColor}{max(min(\Ratio*(\MidNumber - #1)/(\MidNumber-\MinNumber),\Ratio),0.00)} %
            \edef\x{\noexpand\cellcolor{green!\PercentColor!yellow!80}}\x #1
        \fi
}
\newcommand*{\MinN}{4.0}
\newcommand*{\MidN}{6.5} 
\newcommand*{\MaxN}{30}
\newcommand{\cla}[1]{%
        \ifdim #1 pt > \MidN pt
            \pgfmathsetmacro{\PercentColor}{max(min(\Ratio*(#1 - \MidN)/(\MaxN-\MidN),\Ratio),0.00)} %
            \edef\x{\noexpand\cellcolor{red!\PercentColor!yellow!80}}\x #1
        \else
            \pgfmathsetmacro{\PercentColor}{max(min(\Ratio*(\MidN- #1)/(\MidN-\MinN),\Ratio),0.00)} %
            \edef\x{\noexpand\cellcolor{green!\PercentColor!yellow!80}}\x #1
        \fi
}
\begin{document}
\pagestyle{headings}
\mainmatter

\title{PS-FCN: A Flexible Learning Framework for Photometric Stereo} 

\titlerunning{PS-FCN: A Flexible Learning Framework for Photometric Stereo}

\authorrunning{G. Chen, K. Han and K.-Y. K. Wong}

\author{Guanying Chen$^1$ \enspace Kai Han$^2$ \enspace Kwan-Yee K. Wong$^1$}


\institute{
	The University of Hong Kong\\
	\email{ \{gychen,kykwong\}@cs.hku.hk}
    \and
	University of Oxford\\
	\email{khan@robots.ox.ac.uk}
}
\maketitle

\begin{abstract}
This paper addresses the problem of photometric stereo for non-Lambertian surfaces. Existing approaches often adopt simplified reflectance models to make the problem more tractable, but this greatly hinders their applications on real-world objects. In this paper, we propose a deep fully convolutional network, called PS-FCN, that takes an arbitrary number of images of a static object captured under different light directions with a fixed camera as input, and predicts a normal map of the object in a fast feed-forward pass. Unlike the recently proposed learning based method, PS-FCN does not require a pre-defined set of light directions during training and testing, and can handle multiple images and light directions in an order-agnostic manner. Although we train PS-FCN on synthetic data, it can generalize well on real datasets. We further show that PS-FCN can be easily extended to handle the problem of uncalibrated photometric stereo.
Extensive experiments on public real datasets show that PS-FCN outperforms existing approaches in calibrated photometric stereo, and promising results are achieved in uncalibrated scenario, clearly demonstrating its effectiveness.

\keywords{photometric stereo, convolutional neural network}
\end{abstract}

\section{Introduction}

\begin{figure}[t]
\centering
    \includegraphics[width=0.9\textwidth]{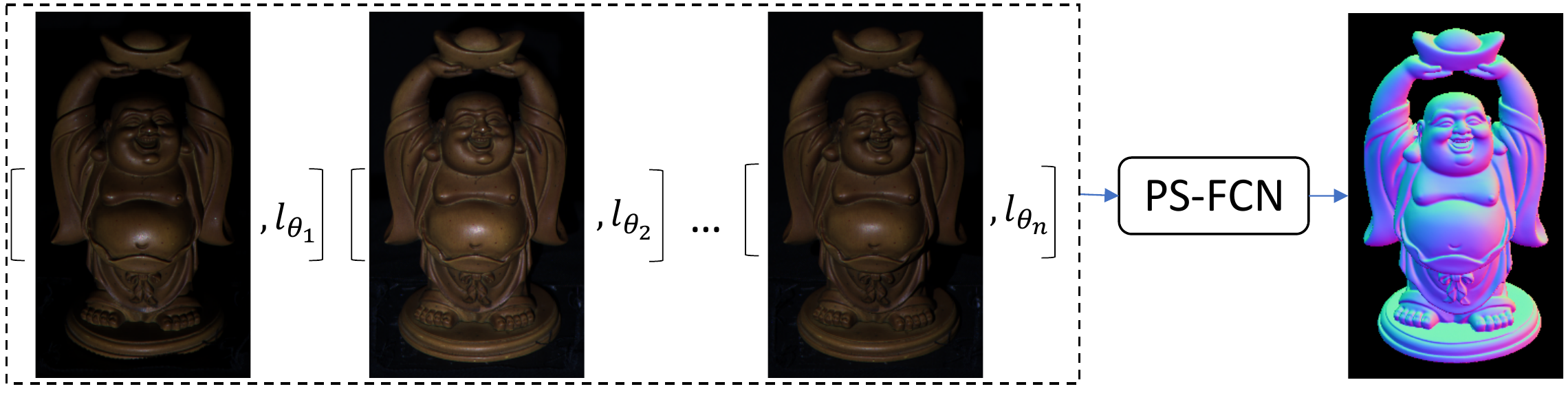}
    \caption{Given an arbitrary number of images and their associated light directions as input, our model estimates a normal map of the object in a fast feed-forward pass.} \label{fig:Intro}
\end{figure}

Given multiple images of a static object captured under different light directions with a fixed camera, the surface normals of the object can be estimated using photometric stereo techniques. Early photometric stereo algorithms often assumed an ideal Lambertian reflectance model~\cite{woodham1980ps,silver1980determining}. Unfortunately, most of the real-world objects are non-Lambertian, and therefore more general models are needed to make photometric stereo methods more practical. Bidirectional reflectance distribution function (BRDF) is a general form for describing the reflectance property of a surface. However, it is difficult to handle general non-parametric BRDFs in non-Lambertian photometric stereo. Many researchers therefore adopted analytical reflectance models~\cite{georghiades2003incorporating,chung2008efficient,ruiters2009heightfield} to simplify the problem. However, a specific analytical model is only valid for a small set of materials. Besides, fitting an analytical model to all the captured data requires solving a complex optimization problem. Hence, it remains an open and challenging problem to develop a computationally efficient photometric stereo method that can handle materials with diverse BRDFs. 

Deep learning frameworks \cite{krizhevsky2012imagenet,lecun1998gradient} have shown great success in both high-level and low-level computer vision tasks. In the context of photometric stereo, Santo \textit{et al.} \cite{santo2017deep} recently proposed a deep fully-connected network, called DPSN, to learn the mapping between reflectance observations and surface normals in a per-pixel manner. For each pixel, DPSN takes observations under 96 pre-defined light directions as input and predicts a normal vector. 
Note that since DPSN depends on a pre-defined set of light directions during training and testing, its practical use is sort of limited. Besides, DPSN predicts a normal vector based solely on the reflectance observations of a single pixel, it cannot take full advantage of the information embedded in the neighborhood of a surface point.

In this paper, we propose a flexible fully convolutional network \cite{long2015fully}, called PS-FCN, for estimating a normal map of an object (see Fig. \ref{fig:Intro}). Convolutional network inherently takes observations in a neighborhood into account in computing the feature map, making it possible for PS-FCN to take advantage of local context information (e.g., surface smoothness prior). PS-FCN is composed of three components, namely a {\em shared-weight feature extractor} for extracting feature representations from the input images, a {\em fusion layer} for aggregating features from multiple input images, and a {\em normal regression network} for inferring the normal map (see Fig.~\ref{fig:network}).

Unlike \cite{santo2017deep}, PS-FCN does not depend on a pre-defined set of light directions during training and testing, and allows the light directions used in testing different from those used in training. It takes an arbitrary number of images with their associated light directions as input, and predicts a normal map of the object in a fast feed-forward pass. It can handle multiple images and light directions in an order-agnostic manner. To simulate real-world complex non-Lambertian surfaces for training PS-FCN, we create two synthetic datasets using shapes from the blobby shape dataset \cite{johnson2011shape} and the sculpture shape dataset \cite{wiles2017silnet}, and BRDFs from the MERL BRDF dataset \cite{matusik2003merl}. After training on synthetic data, we show that PS-FCN can generalize well on real datasets, including the DiLiGenT benchmark \cite{shi2018benchmark}, the Gourd\&Apple dataset \cite{alldrin2008p}, and the Light Stage Data Gallery \cite{einarsson2006relighting}. We further demonstrate that PS-FCN can be easily extended to handle the problem of uncalibrated photometric stereo, which reiterates the flexibility of our model. 
Extensive experiments on public real datasets show that PS-FCN outperforms existing approaches in calibrated photometric stereo, and promising results are achieved in uncalibrated scenario, clearly demonstrating its effectiveness.

\section{Related work}
In this section, we briefly review representative non-Lambertian photometric stereo techniques. More comprehensive surveys of photometric stereo algorithms can be found in \cite{herbort2011introduction,shi2018benchmark}. Non-Lambertian photometric stereo methods can be broadly divided into four categories, namely outlier rejection based methods, sophisticated reflectance model based methods, exemplar based methods, and learning based methods.

Outlier rejection based methods assume non-Lambertian observations to be local and sparse such that they can be treated as outliers. Various outlier rejection methods have been proposed based on rank minimization~\cite{wu2010robust}, RANSAC~\cite{mukaigawa2007analysis}, taking median values~\cite{miyazaki2010median}, expectation maximization~\cite{wu2010photometric}, sparse Bayesian regression~\cite{ikehata2012robust}, etc. Outlier rejection methods generally require lots of input images and have difficulty in handling objects with dense non-Lambertian observations (e.g., materials with broad and soft specular highlights).

Many sophisticated reflectance models have been proposed to approximate the non-Lambertian model, including analytical models like Torrance-Sparrow model~\cite{georghiades2003incorporating}, Ward model~\cite{chung2008efficient}, Cook-Torrance model~\cite{ruiters2009heightfield}, etc.
Instead of rejecting specular observations as outliers, sophisticated reflectance model based methods fit an analytical model to all observations.  These methods require solving complex optimization problems, and can only handle limited classes of materials. Recently, bivariate BRDF representations \cite{shi2014bi,ikehata2014p} were adopted to approximate isotropic BRDF, and a symmetry-based approach \cite{holroyd2008photometric} was proposed to handle anisotropic reflectance without explicitly estimating a reflectance model. 

Exemplar based methods usually require the observation of an additional reference object. Using a reference sphere, Hertzmann and Seitz \cite{hertzmann2005example} subtly transformed the non-Lambertian photometric stereo problem to a point matching problem. Exemplar based methods can deal with objects with spatially-varying BRDFs without knowing the light directions, but the requirement of known shape and material of the reference object(s) limits their applications. As an extension, Hui and Sankaranarayanan \cite{zhui2015ps} introduced a BRDF dictionary to render virtual spheres without using a real reference object, but at the cost of requiring light calibration and longer processing time.

Recently, Santo \textit{et al.} \cite{santo2017deep} proposed a deep fully-connected network, called DPSN, to regress per-pixel normal given a fixed number of observations (e.g., 96) captured under a pre-defined set of light directions. For each image point of the object, all its observations are concatenated to form a fixed-length vector, which is fed into a fully-connected network to regress a single normal vector. DPSN can handle diverse BRDFs without solving a complex optimization problem or requiring any reference objects.
However, it requires a pre-defined set of light directions during training and testing, which limits its practical uses. In contrast, our PS-FCN does not depend on a pre-defined set of light directions during training and testing, and allows the light directions used in testing to be different from those used in training. It takes an arbitrary number of images with their light directions as input, and predicts a normal map of the object in a fast feed-forward pass. It can handle multiple images and light directions in an order-agnostic manner.

Typically, photometric stereo methods require calibrated light directions, and the calibration process is often very tedious. A few works have been devoted to handle uncalibrated photometric stereo (e.g., \cite{alldrin2007r,shi2010self,wu2013calib,papad14closed,lu2015intensity,lu2018symps}). These methods can infer surface normals in the absence of calibrated light directions. Our PS-FCN can be easily extended to handle uncalibrated photometric stereo, by simply removing the light directions during training. Afterwards, it can solely rely on the input images without known light directions to predict the normal map of an object.

\section{Problem formulation}
In this paper, we follow the conventional practice by assuming orthographic projection, directional lights, and the viewing direction pointing towards the viewer. 
Given $q$ color images of an object with $p$ pixels captured under different light directions\footnote{Images are normalized by light intensities, and each light direction is represented by a unit 3-vector.}, a normal matrix $\mathbf{N}_{3\times p}$, a light direction matrix $\mathbf{L}_{3\times q}$, and an observation matrix $\mathbf{I}_{3\times p\times q}$ can be constructed. We further denote the BRDFs for all observations as $\boldsymbol{\Theta}_{3\times p\times q}$, where each 3-vector $\boldsymbol{\Theta}_{:, i, j}$ is a function of the normal, light direction, and viewing direction at $(i, j)$. 
The image formation equation can be written as
\begin{equation}
    \label{eq:eq1}
    \textbf{I} = \boldsymbol{\Theta} \circ \text{repmat}(\textbf{N}^\top \textbf{L}, 3),
\end{equation}
where $\circ$ represents element-wise multiplication, and $\text{repmat}(\mathbf{X},3)$ repeats the matrix $\mathbf{X}$ three times along the first dimension. 

For a Lambertian surface, the BRDF for a surface point degenerates to an unknown constant vector. Theoretically, with three or more independent observations, the albedo scaled surface normal can be solved using linear least squares \cite{woodham1980ps}. However, pure Lambertian surfaces barely exist. We therefore have to consider a more complex problem of non-Lambertian photometric stereo, in which we estimate the normal matrix $\mathbf{N}$ from an observation matrix $\mathbf{I}$ and light direction matrix $\mathbf{L}$ under unknown general BRDFs $\boldsymbol{\Theta}$. 

We design a learning framework based on (\ref{eq:eq1}) to tackle the problem of non-Lambertian photometric stereo. Different from previous methods which approximate $\boldsymbol\Theta$ with some sophisticated reflectance models, our method directly learns the mapping from $(\textbf{I}, \textbf{L})$ to $\textbf{N}$ without explicitly modeling $\boldsymbol\Theta$. 
\section{Learning photometric stereo}
In this section, we first introduce our strategy for adapting CNNs to handle a variable number of inputs, and then present a flexible fully convolutional network, called PS-FCN, for learning photometric stereo. 
\subsection{Max-pooling for multi-feature fusion}
CNNs have been successfully applied to dense regression problems like depth estimation \cite{eigen2014depth} and surface normal estimation \cite{wang2015designing}, where the number of input images is fixed and identical during training and testing. Note that adapting CNNs to handle a variable number of inputs during testing is not straightforward, as convolutional layers require the input to have a fixed number of channels during training and testing. Given a variable number of inputs, a shared-weight feature extractor can be used to extract features from each of the inputs (e.g., siamese networks), but an additional fusion layer is required to aggregate such features into a representation with a fixed number of channels. A convolutional layer is applicable for multi-feature fusion only when the number of inputs is fixed. Unfortunately, this is not practical for photometric stereo where the number of inputs often varies.

One possible way to tackle a variable number of inputs is to arrange the inputs sequentially and adopt a recurrent neural network (RNN) to fuse them. For example, \cite{choy20163d} introduced a RNN framework to unify single- and multi-image 3D voxel prediction. The memory mechanism of RNN enables it to handle sequential inputs, but at the same time also makes it sensitive to the order of inputs. This order sensitive characteristic is not desirable for photometric stereo as it will restrict the illumination changes to follow a specific pattern, making the model less general.

\begin{figure} \centering
    \includegraphics[width=0.9\textwidth]{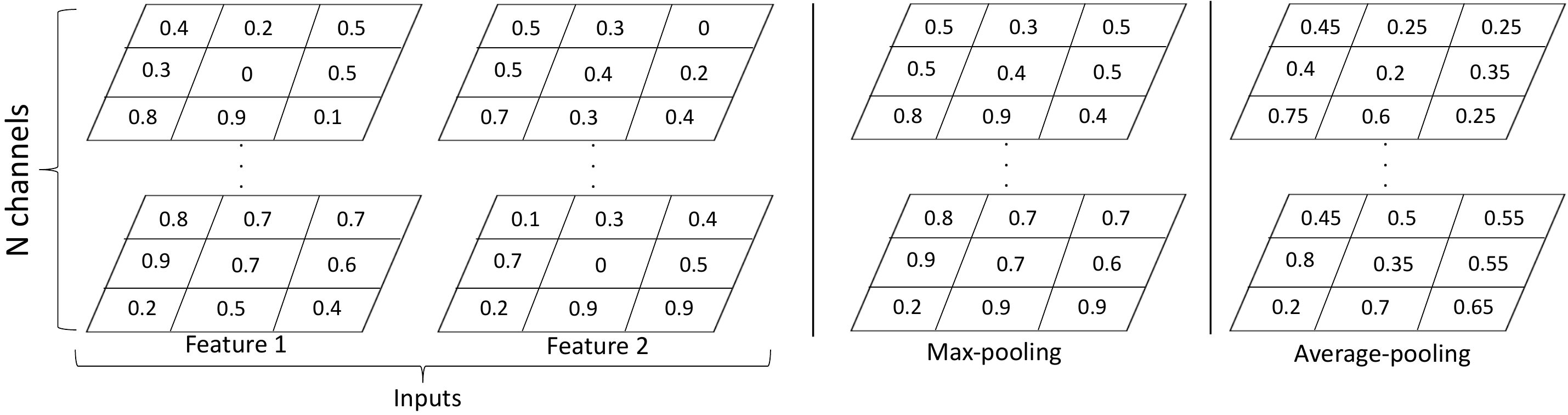}
    \caption{A toy example for max-pooling and average-pooling mechanisms on multi-feature fusion.} \label{fig:pooling}
\end{figure}

More recently, order-agnostic operations (e.g., pooling layers) have been exploited in CNNs to aggregate multi-image information. Wiles and Zisserman \cite{wiles2017silnet} used max-pooling to fuse features of silhouettes from different views for novel view synthesis and 3D voxel prediction. Hartmann \textit{et al.} \cite{hartmann2017learned} adopted average-pooling to aggregate features of multiple patches for learning multi-patch similarity. In general, max-pooling operation can extract the most salient information from all the features, while average-pooling can smooth out the salient and non-activated features. Fig.~\ref{fig:pooling} illustrates how max-pooling and average-pooling operations aggregate two features with a toy example.  

For photometric stereo, we argue that max-pooling is a better choice for aggregating features from multiple inputs. 
Our motivation is that, under a certain light direction, regions with high intensities or specular highlights provide strong clues for surface normal inference (e.g., for a surface point with a sharp specular highlight, its normal is close to the bisector of the viewing and light directions). Max-pooling can naturally aggregate such strong features from images captured under different light directions. Besides, max-pooling can ignore non-activated features during training, making it robust to cast shadow. As will be seen in Section~\ref{sec:exp}, our experimental results do validate our arguments. We observe from experiments that each channel of the feature map fused by max-pooling is highly correlated to the response of the surface to a certain light direction. Strong responses in each channel are found in regions with surface normals having similar directions. The feature map can therefore be interpreted as a decomposition of the images under different light directions (see Fig.~\ref{fig:res_visual}).

\subsection{Network architecture}
\begin{figure}[t] \centering
    \includegraphics[width=0.9\textwidth]{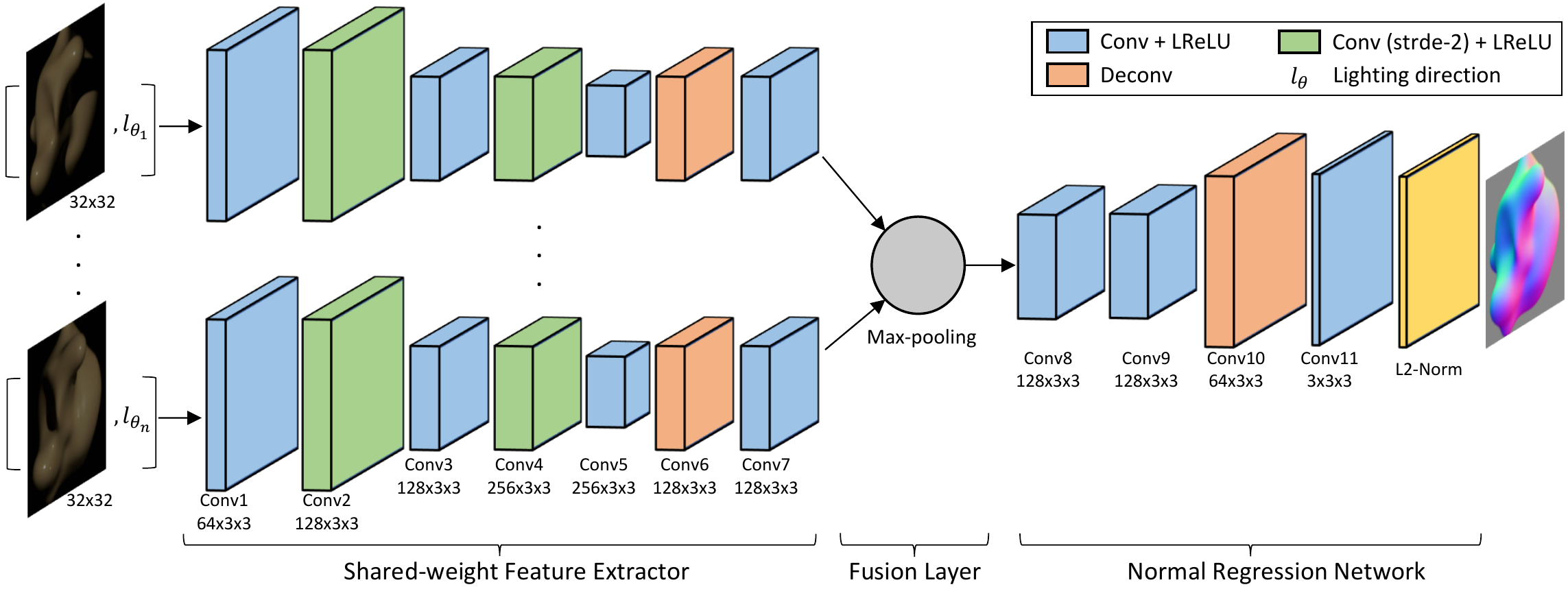}
    \caption{Network architecture of PS-FCN.} \label{fig:network}
\end{figure}

PS-FCN is a multi-branch siamese network \cite{bromley1993signature} consisting of three components, namely a {\em shared-weight feature extractor}, a {\em fusion layer}, and a {\em normal regression network} (see Fig. \ref{fig:network}). It can be trained and tested using an arbitrary number of images with their associated light directions as input.
\label{sub:Network Architecture}

For an object captured under $q$ distinct light directions, we repeat each light direction (i.e., a 3-vector) to form a 3-channel image having the same spatial dimension as the input image ($3\times h\times w$), and concatenate it with the input image. Hence, the input to our model has a dimension of $q\times 6\times h\times w$. 
We separately feed the image-light pairs to the shared-weight feature extractor to extract a feature map from each of the inputs, and apply a max-pooling operation in the fusion layer to aggregate these feature maps. Finally, the normal regression network takes the fused feature map as input and estimates a normal map of the object.

The shared-weight feature extractor has seven convolutional layers, where the feature map is down-sampled twice and then up-sampled once, resulting in a down-sample factor of two. This design can increase the receptive field and preserve spatial information with a small memory consumption. 
The normal regression network has four convolutional layers and up-samples the fused feature map to the same spatial dimension as the input images. An L2-normalization layer is appended at the end of the normal regression network to produce the normal map.

PS-FCN is a fully convolutional network, and it can be applied to datasets with different image scales. Thanks to the max-pooling operation in the fusion layer, it possesses the order-agnostic property. Besides, PS-FCN can be easily extended to handle uncalibrated photometric stereo, where the light directions are not known, by simply removing the light directions during training.

\subsection{Loss function}
\label{sub:Loss function}
The learning of our PS-FCN is supervised by the estimation error between the predicted and the ground-truth normal maps. We formulate our loss function using the commonly used cosine similarity loss
\begin{align}
    \label{eq:cosine}
    L_{normal} = \frac{1}{hw} \sum_{i,j} (1 - \mathbf{N}_{ij} \cdot \tilde{\mathbf{N}}_{ij}),
\end{align}
where $\mathbf{N}_{ij}$ and $\tilde{\mathbf{N}}_{ij}$ denote the predicted normal and the ground truth, respectively,  at the point $(i,j)$. If the predicted normal has a similar orientation as the ground truth, the dot-product $\mathbf{N}_{ij} \cdot \tilde{\mathbf{N}}_{ij}$ will be close to $1$ and the loss will be small, and vice versa. Other losses like mean square error can also be adopted.

\section{Dataset}
The training of PS-FCN requires the ground-truth normal maps of the objects. However, obtaining ground-truth normal maps of real objects is a difficult and time-consuming task. Hence, we create two synthetic datasets for training and one synthetic dataset for testing. The publicly available real photometric stereo datasets are reserved to validate the generalization ability of our model. Experimental results show that our PS-FCN trained on the synthetic datasets generalizes well on the challenging real datasets.

\subsection{Synthetic data for training}
We used shapes from two existing 3D datasets, namely the blobby shape dataset \cite{johnson2011shape} and the sculpture shape dataset \cite{wiles2017silnet}, to generate our training data using the physically based raytracer Mitsuba \cite{jakob2010mitsuba}. Following DPSN \cite{santo2017deep}, we employed the MERL dataset \cite{matusik2003merl}, which contains 100 different BRDFs of real-world materials, to define a diverse set of surface materials for rendering these shapes. 
Note that our datasets explicitly consider cast shadows during rendering.
For the sake of data loading efficiency, we stored our training data in 8-bit PNG format.

\begin{figure} \centering
    \input{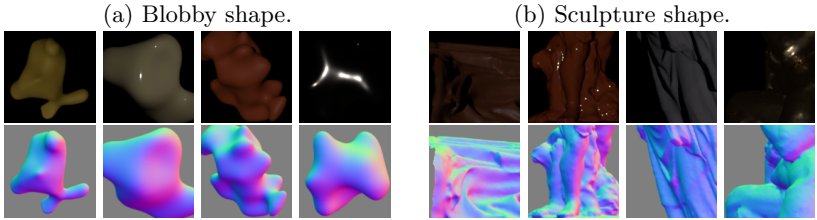}
    \caption{Examples of the synthetic training data.}
    \label{fig:data_samples}
\end{figure}
\noindent{\bf Blobby dataset\enspace} We first followed \cite{santo2017deep} to render our training data using the blobby shape dataset \cite{johnson2011shape}, which contains 10 blobby shapes with various normal distributions. For each blobby shape, $1,296$ regularly-sampled views (36 azimuth angles $\times$ 36 elevation angles) were used, and for each view, 2 out of 100 BRDFs were randomly selected, leading to 25,920 samples ($10\times 36\times 36\times 2$).
For each sample, we rendered 64 images with a spatial resolution of $128 \times 128$ under light directions randomly sampled from a range of 180\degree $\times$ 180\degree, which is more general than the range (74.6\degree $\times$ 51.4\degree) used in the real data benchmark \cite{shi2018benchmark}. We randomly split this dataset into $99:1$ for training and validation (see Fig. \ref{fig:data_samples}(a)).

\noindent{\bf Sculpture dataset\enspace} The surfaces in the blobby shape dataset are usually largely smooth and lack of details. To provide more complex (realistic) normal distributions for training, we employed 8 complicated 3D models from the sculpture shape dataset introduced in \cite{wiles2017silnet}. We generated samples for the sculpture dataset in exactly the same way we did for the blobby shape dataset, except that  we discarded views containing holes or showing uniform normals (e.g., flat facets). 
The rendered images are with a size of $512 \times 512$ when a whole sculpture shape is in the field of view.
We then regularly cropped patches of size $128 \times 128$ from the rendered images and discarded those with a foreground ratio less than 50\%.
This gave us a dataset of 59,292 samples, where each sample contains 64 images rendered under different light directions. Finally, we randomly split this dataset into $99:1$ for training and validation (see Fig. \ref{fig:data_samples}(b)).

\noindent{\bf Data augmentation\enspace} To narrow the gap between real and synthetic data, data augmentation was carried out on-the-fly during training. Given an image of size $128\times 128$, we randomly performed image rescaling (with the rescaled width and height within the range of [32, 128], without preserving the original aspect ratio) and noise perturbation (in a range of [-0.05, 0.05]). Image patches of size $32\times 32$ were then randomly cropped for training. 

\subsection{Synthetic data for testing}
To quantitatively evaluate the performance of our model on different materials and shapes, we rendered a synthetic test dataset using a \textit{Sphere} shape and a \textit{Bunny} shape. Each shape was rendered with all of the 100 BRDFs from MERL dataset under 100 randomly sampled light directions. Similarly, the light directions were sampled from a range of 180\degree $\times$ 180\degree. As a result, we obtained 200 testing samples, and each sample contains 100 images. 

\subsection{Real data for testing}
We employed three challenging real non-Lambertian photometric stereo datasets for testing, namely the DiLiGenT benchmark \cite{shi2018benchmark}, the Gourd\&Apple dataset\cite{alldrin2008p}, and the Light Stage Data Gallery \cite{einarsson2006relighting}. Note that none of these datasets were used during training.

The DiLiGenT benchmark \cite{shi2018benchmark} contains 10 objects of various shapes with complex materials. For each object, 96 images captured under different pre-defined light directions and its ground-truth normal map are provided. We quantitatively evaluated our model on both the main and test datasets of this benchmark. 

The Gourd\&Apple dataset \cite{alldrin2008p} and the Light Stage Data Gallery \cite{einarsson2006relighting} are two other challenging datasets that without ground-truth normal maps. The Gourd\&Apple dataset is composed of three objects, namely \textit{Gourd1}, \textit{Gourd2}, and \textit{Apple}. They provide 102, 98 and 112 image-light pairs, respectively. The Light Stage Data Gallery \cite{einarsson2006relighting} is composed of six objects, and 253 image-light pairs are provided for each object.\footnote{In our experiment, for each object in the Light Stage Data Gallery, we only used the 133 pairs with the front side of the object under illumination.} We qualitatively evaluated our model on these two datasets to further demonstrate the transferability of our model.

\section{Experimental evaluation}
\label{sec:exp}
In this section, we present experimental results and analysis. We carried out network analysis for PS-FCN on the synthetic test dataset, and compared our method with the previous state-of-the-art methods on the DiLiGenT benchmark \cite{shi2018benchmark}. Mean angular error (MAE) in degree was used to measure the accuracy of the predicted normal maps. We further provided qualitative results on the Gourd\&Apple dataset \cite{alldrin2008p} and the Light Stage Data Gallery \cite{einarsson2006relighting}. 

\subsection{Implementation details}
Our framework was implemented in PyTorch \cite{paszke2017pytorch} with 2.2 million learnable parameters. 
We trained our model using a batch size of 32 for 30 epochs, and it only took a few hours for training to converge using a single NVIDIA Titan X Pascal GPU (e.g.,  about 1 hour for 8 image-light pairs per sample on the blobby dataset, and about 9 hours for 32 image-light pairs per sample on both the blobby and sculpture datasets).
Adam optimizer \cite{kingma2014adam} was used with default parameters ($\beta_1=0.9$ and $\beta_2=0.999$), where the learning rate was initially set to 0.001 and divided by 2 every 5 epochs. 
Our code, model and datasets are available at \url{https://guanyingc.github.io/PS-FCN}. 
\begin{table}[t] \centering
    \caption{Results of network analysis on the synthetic test dataset. The numbers represent the average MAE of all the objects (the lower the better). B and S stand for the blobby and sculpture training datasets respectively. ($\dag$ indicates the number of per-sample image-light pairs used is identical during training and testing.)} \label{tab:ablation}
    \setlength{\tabcolsep}{5pt}
\resizebox{0.9\textwidth}{!}{
\begin{tabular}{cccc*{5}{c}}
    \toprule
    \multicolumn{4}{c}{Variants} & \multicolumn{5}{c}{Tested with \# images} \\
                    ID & Data & Fusion Type & Train \#   &  1         & 8             & 16            & 32            & 100    \\
    \midrule
                     0 & B    & Avg-p & 16   & \cla{38.60} & \cla{8.96}  & \cla{6.70} & \cla{6.13} & \cla{5.61} \\
                     1 & B    & Avg-p & 32   & \cla{45.04} & \cla{10.94} & \cla{7.28} & \cla{6.00} & \cla{5.52} \\
    2 & B    & Conv & $\dag$  & - & - &  \cla{7.09}   & \cla{6.49} & -     \\  
    \midrule
    3 & B    & Max-p & 1   &   \B{\cla{22.47}} & \cla{14.58} & \cla{13.95} & \cla{13.88} & \cla{13.67}  \\
                     4 & B    & Max-p & 8   &   \cla{27.96} & \cla{7.40} & \cla{6.24} & \cla{5.87} & \cla{5.82}  \\
                     5 & B    & Max-p & 16  &   \cla{46.85} & \cla{8.44} & \cla{6.24} & \cla{5.64} & \cla{5.43}  \\
                     6 & B    & Max-p & 32  &   \cla{45.17} & \cla{11.84} & \cla{6.64} & \cla{5.50} & \cla{5.30} \\
    \midrule
                    7 &B + S & Max-p & 8 &   \cla{26.65} & \B{\cla{7.20}} & \cla{6.17} & \cla{5.71} & \cla{5.66}  \\
                    8 &B + S & Max-p & 16 &   \cla{36.07} & \cla{7.71} & \B{\cla{5.94}} & \cla{5.29} & \cla{5.03}  \\
                    9 &B + S & Max-p & 32 &   \cla{51.18} & \cla{9.12} & \cla{6.01} & \B{\cla{4.91}} & \B{\cla{4.55}} \\
    \bottomrule
\end{tabular}
}

\end{table}

\subsection{Network analysis}
We quantitatively analyzed PS-FCN on the synthetic test dataset. In particular, we first validated the effectiveness of max-pooling in multi-feature fusion by comparing it with average-pooling and convolutional layers. We then investigated the influence of per-sample input number during training and testing. Besides, we investigated the influence of the complexity of training data. Last, we evaluated the performance of PS-FCN on different materials. For all the experiments in network analysis, we performed 100 random trials (save for the experiments using all 100 image-light pairs per sample during testing) and reported the average results which are summarized in Tab. \ref{tab:ablation}. 

\begin{figure}[t]
\centering
    \input{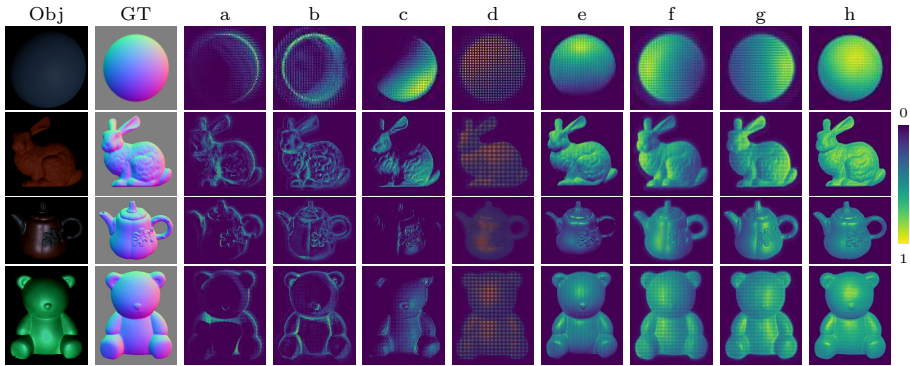}
    \caption{Visualization of the learned feature map after fusion. The first two columns show the images and ground-truth normal maps. Each of the subsequent columns (a-h) shows one particular channel of the fused feature map. 8 out of the 128 channels of the feature map are presented. Note that different regions with similar normal directions are fired in different channels. Each channel can therefore be interpreted as the probability of the normal belonging to a certain direction (or alternatively as the object shading rendered under a certain light direction). Accurate normal maps can then be inferred from these probability distributions.} \label{fig:res_visual}
\end{figure}

\noindent{\bf Effectiveness of max-pooling\enspace} Experiments with IDs 0, 1, 5 \&  6 in Tab.~\ref{tab:ablation} compared the performance of average-pooling and max-pooling for multi-feature fusion. It can be seen that max-pooling performed consistently better than average-pooling, when the per-sample input number during testing was $\geq$ 16. Similarly, experiments with IDs 2, 5 \& 6 showed that fusion by convolutional layers on the concatenated features was sub-optimal. This could be explained by the fact that the weights of the convolutional layers are related to the order of the concatenated features, while the orders of the input image-light pairs are random in our case, thus increasing the difficulty for the convolutional layers to find the relations among multiple features. 
Fig. \ref{fig:res_visual} visualizes the fused features (by max-pooling) of \textit{Sphere} (blue-rubber) \& \textit{Bunny} (dark-red-paint) in synthetic test dataset, and \textit{pot2} \& \textit{bear} in DiLiGenT main dataset. Note that all the image-light pairs were used as input and the features were normalized to $[0, 1]$.

\begin{figure}[h] \centering
    \includegraphics[width=\textwidth]{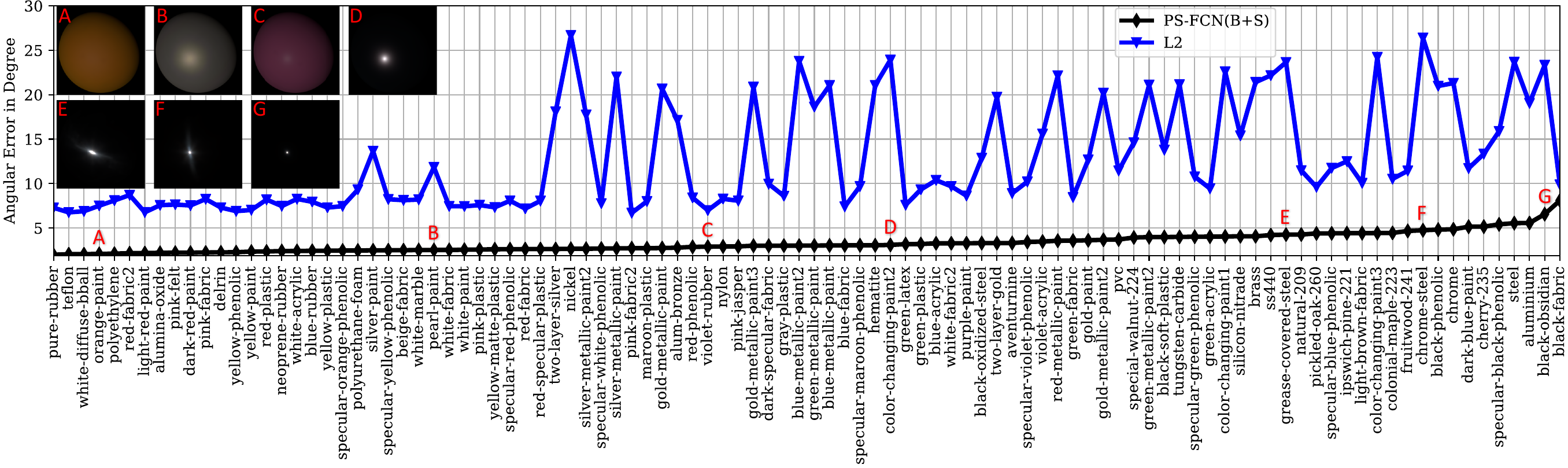}
    \caption{Quantitative comparison between PS-FCN and L2 Baseline \cite{woodham1980ps} on the samples of \textit{Sphere} rendered with 100 different BRDFs. Images in the upper-left corner show the corresponding samples. } \label{fig:100brdfs}
\end{figure}

\noindent{\bf Effects of input number\enspace} Referring to the experiments with IDs $3 - 6$ in Tab. \ref{tab:ablation}, for a fixed number of inputs during training, the performance of PS-FCN increased with the number of inputs during testing. For a fixed number of inputs during testing, PS-FCN performed better when the number of inputs during training was close to that during testing. 

\noindent{\bf Effects of training data\enspace} By comparing experiments with IDs $4 - 6$ (where the models were trained only on the blobby dataset) with experiments with IDs $7 - 9$ (where the models were trained on both the blobby dataset and the sculpture dataset), we can see that the additional sculpture dataset with a more complex normal distribution helped to boost the performance. This suggests that the performance of PS-FCN could be further improved by introducing more complex and realistic training data. 

\noindent{\bf Results on different materials\enspace} Fig. \ref{fig:100brdfs} compares PS-FCN (trained with 32 per-sample inputs on both synthetic datasets) with L2 Baseline \cite{woodham1980ps} on samples of \textit{Sphere} that were rendered with 100 different BRDFs. It can be seen that PS-FCN significantly outperformed L2 Baseline. 
Note that PS-FCN generally performed better on materials with a light color than those with a dark color. This might be explained by the fact that max-pooling always tries to aggregate the most salient features for normal inference, and the image intensities of objects with a dark color are mostly very small. As a result, fewer useful features could be extracted to infer normals for objects of dark materials.

\begin{table}[t]
    \caption{Comparison of results on the DiLiGenT benchmark main dataset. The numbers represent the MAE (the lower the better). Results of PS-FCN under two different testing settings are reported, e.g., PS-FCN (B+S+32, 16) indicates the model trained on both the blobby dataset and the sculpture dataset with a per-sample input number of 32, and tested with a per-sample input number of 16. (Note that the result of PS-FCN (B+S+32, 16) is the average of 100 random trials.)} \label{tab:quant_main}
    \centering
\resizebox{\textwidth}{!}{
    \begin{tabular}{c|*{10}{c}|c}
        \toprule
Method   & ball         &  cat         & pot1         & bear         & pot2         & buddha        & goblet         &reading         &  cow           &harvest        & Avg. \\
        \midrule
L2 \cite{woodham1980ps}   &\cl{4.10}     &\cl{8.41}     &\cl{8.89}     &\cl{8.39}     &\cl{14.65}    &\cl{14.92}    &\cl{18.50}    &\cl{19.80}     &\cl{25.60}    &\cl{30.62}     &\cl{15.39} \\
AZ08 \cite{alldrin2008p}  &\cl{2.71}     &\cl{6.53}     &\cl{7.23}     &\B{\cl{5.96}}     &\cl{11.03}    &\cl{12.54}    &\cl{13.93}    &\cl{14.17}     &\cl{21.48}    &\cl{30.50}     &\cl{12.61} \\
WG10 \cite{wu2010robust}  &\cl{2.06}     &\cl{6.73}     &\cl{7.18}     &\cl{6.50}     &\cl{13.12}    &\cl{10.91}    &\cl{15.70}    &\cl{15.39}     &\cl{25.89}    &\cl{30.01}     &\cl{13.35} \\
IA14 \cite{ikehata2014p}  &\cl{3.34}     &\cl{6.74}     &\cl{6.64}     &\cl{7.11}     &\cl{8.77}     &\cl{10.47}    &\cl{9.71}     &\cl{14.19}     &\cl{13.05}    &\cl{25.95}     &\cl{10.60} \\
ST14 \cite{shi2014bi}     &\B{\cl{1.74}} &\B{\cl{6.12}}     &\B{\cl{6.51}}    &\cl{6.12}     &\cl{8.78}     &\cl{10.60}    &\cl{10.09}    &\cl{13.63}     &\cl{13.93}    &\cl{25.44}     &\cl{10.30} \\  
DPSN \cite{santo2017deep} &\cl{2.02}     &\cl{6.54}     &\cl{7.05}     &\cl{6.31}     &\cl{7.86}     &\cl{12.68}    &\cl{11.28}    &\cl{15.51}     &\cl{8.01}     &\cl{16.86}     &\cl{9.41} \\
        \midrule
        PS-FCN (B+S+32, 16)  & \cl{3.31}	 &\cl{7.64}	 &\cl{8.14}	 &\cl{7.47}	 &\cl{8.22}	 &\cl{8.76}	 &\cl{9.81}	 &\cl{14.09}	 &\cl{8.78}	 &\cl{17.48}	 &\cl{9.37} \\
PS-FCN (B+S+32, 96)  &\cl{2.82}	 &\cl{6.16}	 &\cl{7.13}	 &\cl{7.55}	 &\B{\cl{7.25}}	 &\B{\cl{7.91}}	 &\B{\cl{8.60}}	 &\B{\cl{13.33}}	 &\B{\cl{7.33}}	 &\B{\cl{15.85}}	 &\B{\cl{8.39}} \\
        \bottomrule                                                                     
    \end{tabular}
}

\end{table}

\begin{figure}[h!] \centering
    \input{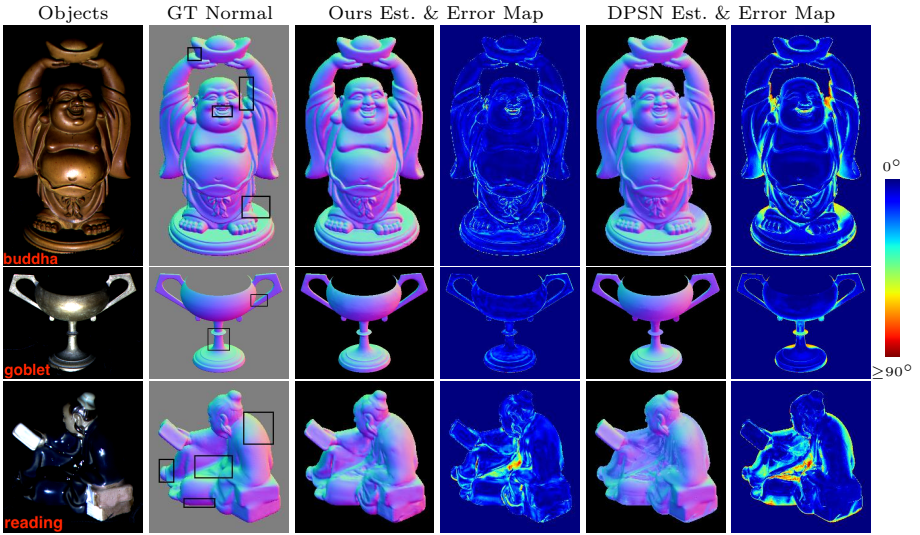}
    \caption{Qualitative results on the DiLiGenT benchmark main dataset. The black boxes in the ground-truth normal maps are regions with cast shadows. Our method can produce more robust estimations in those regions compared with DPSN \cite{santo2017deep}.} \label{fig:qual_main}
\end{figure}

\subsection{Benchmark comparisons}
\label{sub:Results on Real Data}

\noindent{\bf DiLiGenT benchmark main dataset\enspace}
We compared PS-FCN against the recently proposed learning based method DPSN \cite{santo2017deep} and other previous state-of-the-art methods. Quantitative results on the main dataset of the DiLiGenT benchmark are shown in Tab. \ref{tab:quant_main}. 
Compared with other methods, PS-FCN performed particularly well on objects with complicated shapes (e.g., $buddha$, $reading$, and $harvest$) and/or spatially varying materials (e.g., $pot2$, $goblet$ and $cow$). Our best performer, which achieved an average MAE of 8.39$^\circ$, was trained with 32 per-sample inputs on both synthetic datasets and tested with all 96 inputs for each object. With only 16 inputs per object during testing, PS-FCN still outperformed the previous methods in terms of the average MAE. Fig. \ref{fig:qual_main} presents the qualitative comparison between PS-FCN and DPSN. It can be seen that PS-FCN is more robust in regions with cast shadows. 

\noindent{\bf DiLiGenT benchmark test dataset\enspace}
We further evaluated our model on the test dataset of the DiLiGenT benchmark, with the ground-truth normal maps withheld by the original authors (see. Tab. \ref{tab:quant_test}). Similar to the results on the main dataset, PS-FCN outperformed other methods on the test dataset. More results of the other methods can be found on the benchmark website\footnote{\scriptsize\url{https://sites.google.com/site/photometricstereodata/home/summary-of-benchmarking-results}} for comparison. 

\begin{table}[t]
    \caption{Comparison of results on the DiLiGenT benchmark test dataset. The numbers represent the MAE (the lower the better).} \label{tab:quant_test}
    \centering
\resizebox{1.0\textwidth}{!}{
    \begin{tabular}{c|*{9}{c}|c}
        \toprule
Method                    &  cat         & pot1         & bear         & pot2         & buddha        & goblet         &reading         &  cow           &harvest        & Avg. \\
        \midrule                                                                        
        IA14 \cite{ikehata2014p}  &\B{\cl{5.61}} &\B{\cl{6.33}} &\B{\cl{5.12}}     &\cl{8.83}     &\cl{11.00}     &\cl{10.54}      &\B{\cl{13.27}}  &\cl{11.18}      &\cl{24.82}     &\cl{10.74} \\
ST14 \cite{shi2014bi}     &\cl{6.43}     &\cl{6.64}     &\cl{6.09}     &\cl{8.94}     &\cl{10.92}     &\cl{10.33}      &\cl{14.16}      &\cl{10.82}      &\cl{25.43}     &\cl{11.08} \\
DPSN \cite{santo2017deep} &\cl{5.82}     &\cl{8.26}     &\cl{6.32}     &\cl{9.02}     &\cl{12.80}     &\cl{12.04}      &\cl{16.11}      &\cl{8.00}       &\cl{17.78}     &\cl{10.68} \\
        \midrule                                                                       
        PS-FCN (B+S+32, 96) & \cl{6.24} & \cl{7.59} & \cl{5.42} & \B{\cl{7.11}} & \B{\cl{8.30}} & \B{\cl{8.62}} & \cl{13.43} & \B{\cl{7.98}} & \B{\cl{15.93}} & \B{\cl{8.96}}\\ 
        \bottomrule                                                                     
    \end{tabular}
}

\end{table}

\noindent{\bf Uncalibrated photometric stereo extension\enspace}
PS-FCN can be easily extended to handle uncalibrated photometric stereo by simply removing the light directions from the input. To verify the potential of our framework towards uncalibrated photometric stereo, we trained an uncalibrated variant of our model, denoted as UPS-FCN, taking only images as input (note that we assume the images were normalized by the light intensities). UPS-FCN was trained on both synthetic datasets using 32 image-light pairs as input. We compared our UPS-FCN with the existing uncalibrated methods. The results are reported in Tab.~\ref{tab:uncalib}, our UPS-FCN outperformed existing methods in terms of the average MAE, which demonstrates the effectiveness and flexibility of our model.

\begin{table}
    \caption{Comparison of results for uncalibrated photometric stereo on the DiLiGenT benchmark main dataset. The numbers represent the MAE (the lower the better).} \label{tab:uncalib}
    \centering
\resizebox{1.0\textwidth}{!}{
\begin{tabular}{c|*{10}{c}|c}
\toprule
Method                   & ball         & cat          & pot1         & bear         & pot2         & buddha        & goblet         &reading         & cow            &harvest        & Avg. \\
\midrule
AM07 \cite{alldrin2007r} &\cl{7.27}	    &\cl{31.45}	   &\cl{18.37}    &\cl{16.81}	 &\cl{49.16}	 &\cl{32.81}	 &\cl{46.54}	 &\cl{53.65}	  &\cl{54.72}     &\cl{61.70}     &\cl{37.25} \\ 
SM10 \cite{shi2010self}	 &\cl{8.90}	    &\cl{19.84}	   &\cl{16.68}    &\cl{11.98}	 &\cl{50.68}	 &\cl{15.54}	 &\cl{48.79}	 &\cl{26.93}	  &\cl{22.73}     &\cl{73.86}     &\cl{29.59} \\
WT13 \cite{wu2013calib}	 &\B{\cl{4.39}} &\cl{36.55}	   &\B{\cl{9.39}} &\B{\cl{6.42}} &\cl{14.52}	 &\B{\cl{13.19}} &\cl{20.57} &\cl{58.96}	  &\cl{19.75}     &\cl{55.51}     &\cl{23.93} \\
PF14 \cite{papad14closed}&\cl{4.77}	    &\B{\cl{9.54}} &\cl{9.51} 	  &\cl{9.07}	 &\cl{15.90}	 &\cl{14.92}	 &\cl{29.93}	 &\cl{24.18}	  &\cl{19.53}     &\cl{29.21}     &\cl{16.66} \\
LC18 \cite{lu2018symps}  &\cl{9.30}	  &  \cl{12.60} &\cl{12.40} 	  &\cl{10.90}	 &\cl{15.70}	 &\cl{19.00}	 &\B{\cl{18.30}}	 &\B{\cl{22.30}}	  &\cl{15.00}     &\cl{28.00}     &\cl{16.30} \\
\midrule                                                 
UPS-FCN   &\cl{6.62}	 &\cl{14.68}	 &\cl{13.98} &\cl{11.23}	 &\B{\cl{14.19}}	 &\cl{15.87}	 &\cl{20.72}	 &\cl{23.26}	 &\B{\cl{11.91}}	 &\B{\cl{27.79}}  &\B{\cl{16.02}} \\
\bottomrule
\end{tabular}
}

\end{table}

\subsection{Testing on other real datasets}
\begin{figure}[tb]
\centering
    \input{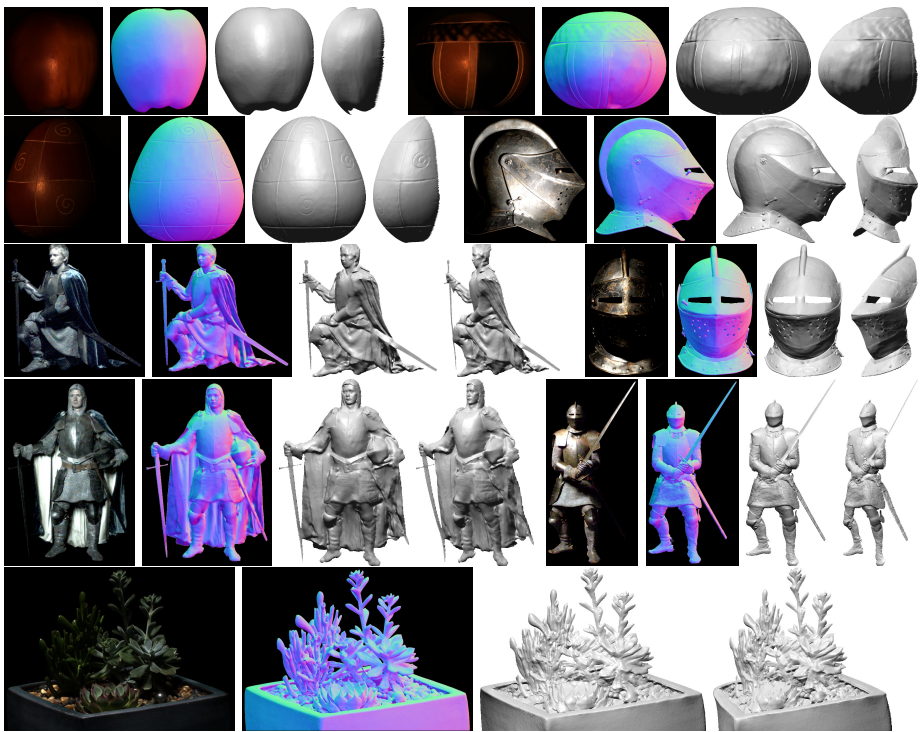}
    \caption{Qualitative results for the Gourd\&Apple dataset and Light Stage Data Gallery. For each shape, a sample input image, the estimated normal map, and two views of the reconstructed surfaces are shown. (Best viewed in PDF with zoom.)} \label{fig:qual_other}
\end{figure}

Due to absence of ground-truth normal maps, we qualitatively evaluated our best-performing model PS-FCN (B+S+32) on the Gourd\&Apple dataset \cite{alldrin2008p} and the Light Stage Data Gallery \cite{einarsson2006relighting}. Fig.~\ref{fig:qual_other} shows the estimated normal maps and surfaces reconstructed using \cite{frankot1988method}. The reconstructed surfaces convincingly reflect the shapes of the objects, demonstrating the accuracy of the normal maps predicted by PS-FCN. 

\section{Conclusions}
In this paper, we have proposed a flexible deep fully convolutional network, called PS-FCN, that accepts an arbitrary number of images and their associated light directions as input and regresses an accurate normal map. Our PS-FCN does not require a pre-defined set of light directions during training and testing, and allows the light directions used in testing different from that used in training. It can handle multiple images and light directions in an order-agnostic manner. In order to train PS-FCN, two synthetic datasets with various realistic shapes and materials have been created. After training, PS-FCN can generalize well on challenging real datasets. 
In addition, PS-FCN can be easily extended to handle uncalibrated photometric stereo. 
Results on diverse real datasets have clearly shown that PS-FCN outperforms previous calibrated photometric stereo methods, and promising results have been achieved in uncalibrated scenario. 

\paragraph{\bf Acknowledgments}
We thank Hiroaki Santo for his help with the comparison to DPSN. We also thank Boxin Shi and Zhipeng Mo for their help with the evaluation on the DiLiGenT benchmark. We gratefully acknowledge the support of NVIDIA Corporation with the donation of the Titan X Pascal GPU used for this research. Kai Han is supported by EPSRC Programme Grant Seebibyte EP/M013774/1.

\bibliographystyle{splncs}
\bibliography{gychen}
\end{document}